\documentclass[10pt,twocolumn,letterpaper]{article}

\usepackage[pagenumbers]{cvpr} % To force page numbers, e.g. for an arXiv version

\definecolor{cvprblue}{rgb}{0.21,0.49,0.74}
\usepackage[pagebackref,breaklinks,colorlinks,allcolors=cvprblue]{hyperref}
\usepackage{arydshln}  % Enables dashed lines in tables
\usepackage[most]{tcolorbox}
\usepackage[table]{xcolor}
\usepackage{wrapfig}
\usepackage{float}

\definecolor{realsonyblue}{RGB}{0, 51, 102}
\definecolor{sonyblue}{RGB}{0, 0, 255}

\newtcolorbox{researchquestion}{
    enhanced,
    colback=white,
    colframe=white,
    borderline west={2pt}{0pt}{realsonyblue},
    left=5pt, right=0pt, top=0pt, bottom=0pt,
    boxrule=0pt,
}

\def\confName{CVPR}
\def\confYear{2027}

\title{VideoNPC: Agentic Navigation with Multimodal Memory and Predictive Control for Long-Video Understanding}

\author{Wei-Yao Wang\thanks{Project Lead: Wei-yao.Wang@sony.com.}~ \quad Kazuya Tateishi \quad Takashi Shibuya \quad Shusuke Takahashi \quad Yuki Mitsufuji\\
Sony Group Corporation\\
{\normalsize\textbf{Project Page : \href{https://wywywang.github.io/VideoNPC-demo}{\texttt{https://wywywang.github.io/VideoNPC-demo/}}}}
}

\newcommand{\modelname}{VideoNPC\xspace}

\begin{document}
\maketitle

\begin{abstract}
Agentic long-video understanding requires an agent to locate sparse, temporally distributed evidence to answer complex questions.
Despite advances, existing video agents commonly rely on text-centric, flat memories and select actions myopically, without evaluating whether an action can reduce uncertainty in subsequent retrieval steps before executing it.
In this paper, we propose \modelname, an agentic \textbf{\underline{N}}avigation framework that includes a multimodal navigational memory with \textbf{\underline{P}}redictive \textbf{\underline{C}}ontrol for long-video understanding.
To organize long videos for relational navigation, the memory consists of a multimodal index that encodes visual streams and text captions into a unified space and a locator-concept-evidence graph that retrieves multifaceted and grounded relations.
In parallel, we introduce model predictive retrieval control (MPRC), which asks the agent to plan candidate proposals and tool-execution actions and then selects one for execution using a verifier that evaluates the expected utility gain estimated by a dynamics model.
Unlike conventional direct reason-then-act pipelines, MPRC anticipates downstream information consequences through receding-horizon planning.
Experiments on EgoSchema, LVBench, VideoMME, and LongVideoBench demonstrate that \modelname achieves relative improvements of 5.4\% over agentic baselines and 7.8\% over its strongest base model, highlighting the benefits of combining multimodal navigational memory with uncertainty-aware receding-horizon planning.
% \footnote{Code will be publicly available in the camera-ready version.}.
\end{abstract}
    
\vspace{-7pt}
\section{Introduction}
\label{sec:intro}

\begin{figure*}[t]
    \centering
    \includegraphics[width=\linewidth]{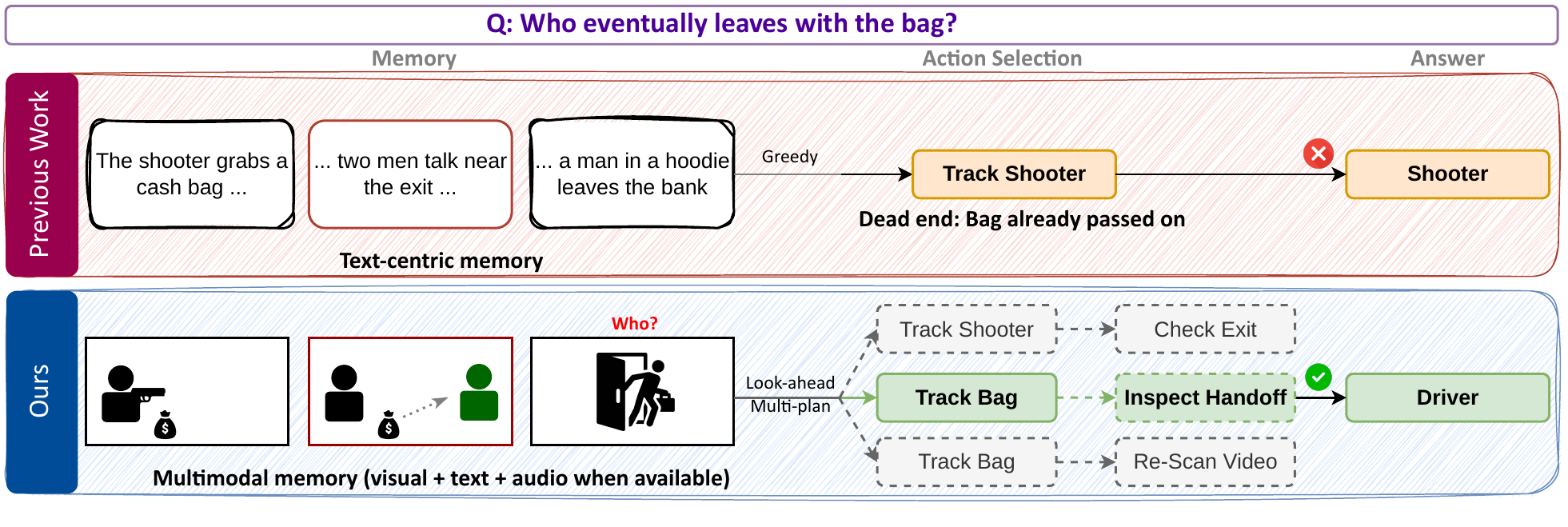}
    \vspace{-28pt}
    \caption{Motivating paradigm comparing existing approaches that rely on text-centric memory and the direct reason-then-act framework (top) with our \modelname, which integrates multimodal navigational memory with model predictive control (bottom).}
    \vspace{-10pt}
    \label{fig:motivation-example}
\end{figure*}

Long-video understanding (LVU) asks a model to answer questions whose supporting evidence may be separated by minutes or hours of video content, serving as a fundamental and practical task for multimodal intelligence in applications such as movies, lectures, and meetings.
Although large multimodal models have made substantial progress on short clips (e.g., with VLMs \cite{DBLP:conf/nips/LiuLWL23a,openai-o3,DBLP:journals/corr/abs-2511-21631}, embedding models \cite{DBLP:journals/corr/abs-2608-27044,DBLP:journals/corr/abs-2510-00523,DBLP:journals/tmlr/MengJLSYFQTZCXXZCY26,DBLP:journals/corr/abs-2604-23321}), naively presenting an entire long video to such a model remains computationally expensive and dilutes the few frames or clips that actually decide the answer \cite{DBLP:conf/nips/MangalamAM23,DBLP:conf/nips/WuLCL24,DBLP:conf/iccv/WangHHCZQDGHXDT25,DBLP:conf/cvpr/0001S0WLXQYXZH025}.
To advance LVU, a growing body of work fine-tunes and post-trains video-LLMs to support reasoning over long-horizon videos \cite{DBLP:conf/emnlp/WangYYIBB25,DBLP:journals/corr/abs-2608-20492,DBLP:conf/eccv/GuanYLJLLLB26}.
In parallel, another line of research explores test-time frameworks in which LLM-based agents iteratively determine where to inspect videos to gather relevant evidence \cite{DBLP:conf/eccv/WangZZY24,DBLP:conf/eccv/FanMWDLGL24,DBLP:conf/cvpr/WangYSYCBB25}. 
We focus on agentic LVU, which leverages the reasoning and planning capabilities of LLMs to enable adaptive, evidence-driven video analysis at test time, without requiring additional model training.

Despite recent progress, most video agents (e.g., \cite{DBLP:conf/nips/ZhangJGLLLL25,DBLP:conf/cvpr/LiLSLCS26,DBLP:conf/cvpr/LinWLSWYLLB26,DBLP:journals/corr/abs-2608-07663}) suffer from two challenges, as shown in Fig. \ref{fig:motivation-example}:
(1) Agentic methods mostly rely on textual representations, such as captions, transcripts, and semantic summaries, to organize and reason about video content.
Although some agents can inspect raw frames or clips through tools, visual evidence is often insufficiently precise to be represented in a form that supports effective matching with the query or integration into the agent's evolving memory.
Therefore, fine-grained visual cues, such as brief nods, object positions, or recurring unlabeled objects, may be overlooked despite being present in the video.
(2) Existing video agents often adopt ReAct-style frameworks \cite{DBLP:conf/iclr/YaoZYDSN023}, which select and directly execute tool calls through stepwise reasoning without explicitly accounting for how each action will reduce uncertainty relevant to subsequent decisions.
For example, given the query ``Who eventually leaves with the bag?", a retrieved clip may show someone picking up the bag without revealing who ultimately leaves with it. The agent could track either the person or the bag across subsequent events. 
Although tracking the initial carrier may seem more direct, it could lead to a dead end if the bag is handed to someone else. 
Therefore, choosing effectively requires anticipating how alternative actions may unfold, comparing their likely downstream observations and ability to establish the answer, and verifying whether the predicted evidence is sufficient to support a conclusion. 
Effective action selection requires evaluating the expected outcomes of multi-step action sequences, instead of choosing the next action based solely on its immediate utility.
These challenges give rise to a central yet underexplored question:
\begin{researchquestion}
\itshape
How can a video agent effectively acquire and integrate complementary textual and visual evidence, together with audio when available, while planning actions based on their expected downstream outcomes and grounding its conclusions in observed video evidence?
\end{researchquestion}

In this paper, we propose \modelname, an agentic navigation framework that combines a multimodal navigational memory with model predictive retrieval control (MPRC). \modelname first addresses the memory challenge by indexing multimodal clips in a unified embedding space and linking them through the proposed locator-concept-evidence (LCE) graph, enabling direct multimodal matching together with associative, temporal, and multimodal-similarity expansion.
To address the myopic control problem, MPRC asks the agent to propose multiple retrieval trajectories and uses a dynamics model to predict their expected information states.
A trajectory verifier then compares these predictions, selects the trajectory expected to contribute the most useful evidence, and executes its first action before replanning from the resulting observation.

In summary, our contributions are three-fold:
\begin{itemize}
    \item \textbf{Rethinking memory composition for video agents.}
    We introduce a multimodal navigational memory that couples a unified multimodal clip index with an evidence-grounded LCE graph. By integrating direct multimodal evidence retrieval with relational navigation across temporally distant observations, our architecture enables complementary evidence acquisition beyond flat similarity retrieval or hierarchical video organization alone.
    \item \textbf{Enabling predictive retrieval planning for agentic LVU.}
    We propose MPRC, an LVU-specific paradigm of receding-horizon planning that anticipates the information consequences of candidate retrieval trajectories. By explicitly modeling expected evidence categories and residual uncertainties, MPRC enables \modelname to select retrieval actions based on their anticipated contributions to question answerability and information gain.
    \item \textbf{Comprehensive evaluation that validates the effectiveness of our method.} \modelname significantly outperforms existing video agents across multiple long-video benchmarks, with improvements of 2.3\%, 1.1\%, 11.8\%, and 6.4\% (5.4\% on average) over agentic methods and 3.4\%, 9.1\%, 7.4\%, and 11.2\% (7.8\% on average) over its base model (GPT-6.1-Sol \cite{openai-gpt61}), on EgoSchema, LVBench, VideoMME, LongVideoBench, respectively. Our systematic ablations reveal the impacts of multimodal navigational memory and of trajectories and actions in MPRC, providing a general principle for LVU agents.
\end{itemize}
\section{Related work}
\label{sec:related-work}

\begin{figure*}[t]
    \centering
    \includegraphics[width=\linewidth]{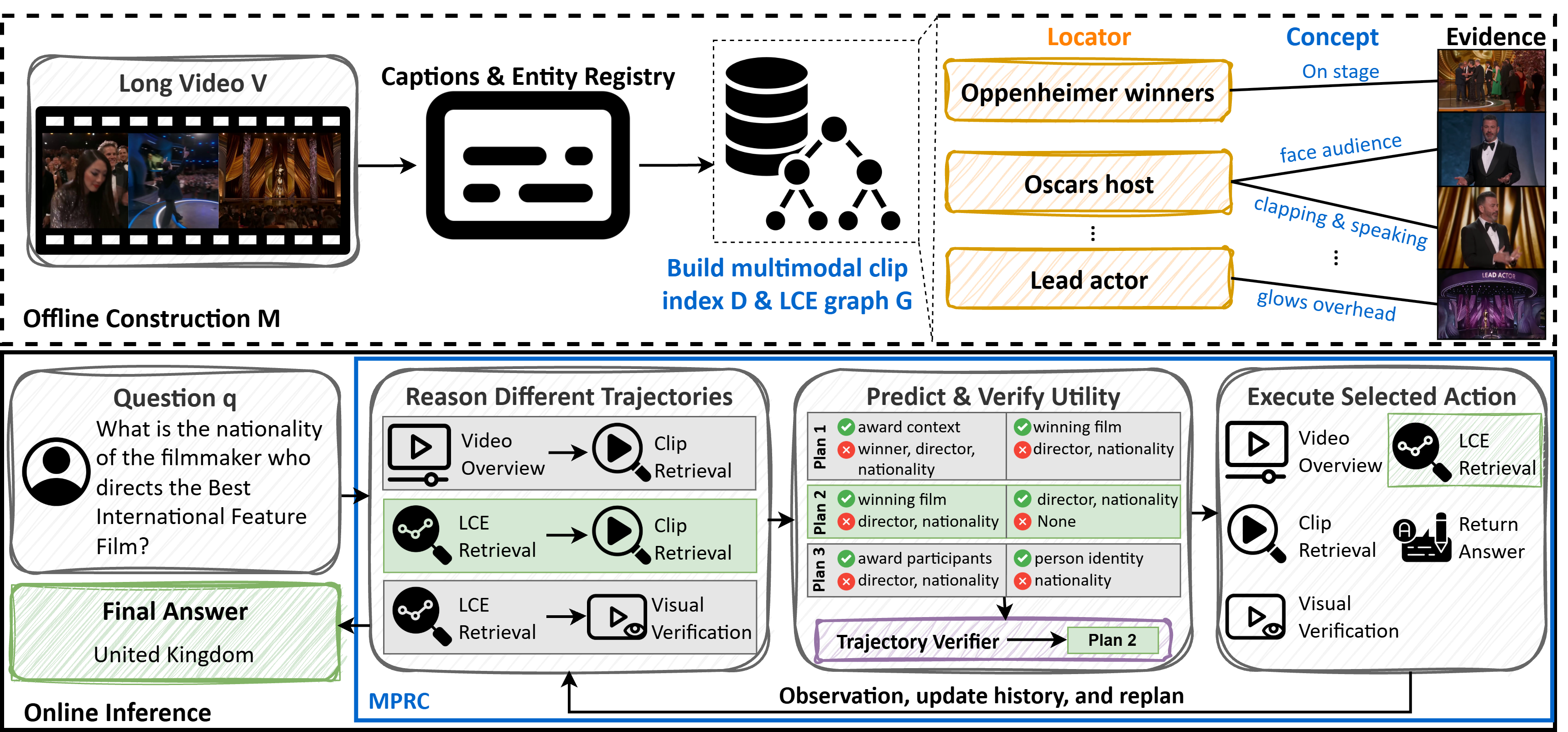}
    \vspace{-17pt}
    \caption{\modelname overview. \textbf{Top:} An offline memory includes a unified multimodal clip index with a Locator-Concept-Evidence (LCE) graph, supporting direct evidence matching and concept-grounded navigation across temporally distant clips. \textbf{Bottom:} At each inference step, MPRC asks the agent to propose candidate retrieval trajectories. A retrieval dynamics model predicts their expected information categories, targeted uncertainties, and residual uncertainties, while a trajectory verifier compares their anticipated utility. Only the first action of the selected trajectory is executed, after which the agent updates its observation history and replans.
    }
    \vspace{-5pt}
    \label{fig:method-overview}
\end{figure*}

\noindent\textbf{Agentic long-video understanding and memory.}
Multimodal LLMs have substantially advanced video understanding \cite{DBLP:journals/corr/abs-2605-25979,DBLP:journals/corr/abs-2511-21631,DBLP:journals/corr/abs-2607-02770,openai-gpt61,openai-o3,DBLP:conf/iclr/Ye0LH0000025,DBLP:journals/corr/abs-2606-12195,DBLP:conf/acl/WangSZWCN25,DBLP:journals/corr/abs-2507-06261}, whereas long-video benchmarks expose the difficulty of locating sparse evidence across temporally extended videos~\cite{DBLP:conf/nips/MangalamAM23,DBLP:conf/nips/WuLCL24,DBLP:conf/iccv/WangHHCZQDGHXDT25,DBLP:conf/cvpr/0001S0WLXQYXZH025,DBLP:conf/cvpr/FuDLLRZWZSZCLLZ25}, motivating agents that adaptively decide where and how densely to inspect. 
VideoAgent~\cite{DBLP:conf/eccv/WangZZY24} iteratively retrieves and captions relevant frames, while ~\citet{DBLP:conf/eccv/FanMWDLGL24} maintain explicit memory through an observe-think-act-memorize loop. VideoTree~\cite{DBLP:conf/cvpr/WangYSYCBB25} organizes evidence into an adaptive coarse-to-fine hierarchy; DVD~\cite{DBLP:conf/nips/ZhangJGLLLL25} designs different tools over a multi-granular video database; and VideoARM~\cite{yin2026videoarm} integrates adaptive video exploration with hierarchical memory updates during reasoning. VideoSeek~\cite{DBLP:conf/cvpr/LinWLSWYLLB26} and LensWalk~\cite{DBLP:conf/cvpr/LiLSLCS26} further adapt the retrieval flow, temporal scope, and observation density to the evolving query. 
However, adaptive video exploration does not necessarily ensure effective retrieval and integration of heterogeneous evidence. 
Hierarchical memories support coarse-to-fine navigation, but their organizational structure alone does not explicitly connect recurring entities and events across temporally distant observations. Meanwhile, memories relying primarily on captions, transcripts, or semantic summaries may overlook fine-grained visual and spatial cues that are difficult to capture in text. Conversely, flat multimodal retrieval enables direct matching to visual evidence but lacks explicit relational navigation across clips. To bridge these complementary limitations, VideoNPC couples a unified multimodal clip index with an LCE graph, enabling direct multimodal evidence matching and relational navigation across temporally distant observations within a unified memory architecture.

\noindent\textbf{Tool use and predictive planning.}
LLMs have enabled autonomous agents through reasoning and planning~\cite{DBLP:journals/inffus/SapkotaRK26}.
For instance, ReAct~\cite{DBLP:conf/iclr/YaoZYDSN023}, widely adopted by video agents, interleaves reasoning with tool execution but selects actions without explicitly anticipating their downstream consequences.
\citet{DBLP:conf/nips/LiGZMMSYWJZL25} improve multimodal tool-use policies through trajectory exploration and step-wise preference tuning, yet remain reactive at inference time.
Beyond video understanding, predictive-decoding~\cite{DBLP:conf/iclr/MaZZHK25} uses simulated trajectories to guide LLM generation in other domains, while \citet{wang2026ampqc} employ receding-horizon planning for visual design.
Despite demonstrating the benefits of non-myopic planning, these approaches do not explicitly address how to evaluate retrieval trajectories by their expected resolution of question-relevant uncertainties in LVU.
To this end, we introduce model predictive retrieval control (MPRC), which adapts receding-horizon planning to predict anticipated results and residual uncertainties, enabling a trajectory-level verifier to compare heterogeneous retrieval plans by their expected contributions to question utility.

\section{Method}
\label{sec:method}

\noindent\textbf{Problem Definition.}
Given a long video $V$ and a question $q$, we formulate long-video understanding as a long-horizon agentic reasoning problem, where an agent iteratively acquires and integrates question-relevant evidence through tool interactions to produce a grounded answer.
At each turn $t$, the agent maintains a decision state $b_t$ and executes a tool action $a_t$ with grounded arguments, receiving an observation $o_t$.
The interaction history $\mathcal{H}_t$ and decision state $b_t$ are defined as:
\begin{equation}
    \mathcal{H}_t =
    \big(
    \langle a_1,o_1\rangle,\ldots,
    \langle a_t,o_t\rangle
    \big), b_t = (q,\mathcal{M},\mathcal{H}_{t-1}),
    \label{eq:interaction-history}
\end{equation}
where $\mathcal{M}$ denotes the video database constructed offline from $V$.
Importantly, $\mathcal{H}_t$ contains only executed interactions and their corresponding tool observations; the estimated candidate plans considered by MPRC (Sec.~\ref{sec:mprc}) are not part of the executed history.

\noindent\textbf{\modelname Overview.}
As shown in Fig.~\ref{fig:method-overview}, \modelname follows an iterative reason-act-observe loop, combining multimodal navigational memory with MPRC to guide evidence acquisition under a maximum interaction budget of $T$ turns.
At each step, the agent reasons over the current decision state $b_t$, proposes alternative tool-use plans, and executes the next action from the most promising trajectory selected by a verifier considering expected utility.
The agent then incorporates the resulting observation into its history $\mathcal{H}_{t}$ before replanning.
Specifically, two components enable this process: (1) Multimodal navigational memory (offline construction), which couples a unified multimodal clip index with a Locator-Concept-Evidence (LCE) graph to provide access to visual evidence and cross-clip relationships (Sec.~\ref{sec:memory}); and (2) MPRC (online inference), which compares the expected utility of different tool-use plans to select the next action (Sec.~\ref{sec:mprc}).
This closed loop allows subsequent actions to adapt to newly retrieved evidence rather than follow a fixed retrieval sequence.

\subsection{Multimodal Navigational Memory}
\label{sec:memory}
\noindent\textbf{Navigational database construction.}
While DVD~\cite{DBLP:conf/nips/ZhangJGLLLL25} introduces a multi-granular database for LVU, its text-only representation limits access to visual evidence, while flat retrieval makes it ineffective to establish relationships across distant clips. 
To address these limitations, we aim to construct a persistent video database: $\mathcal{M}(V)=\big(\mathcal{D},\mathcal{G}\big)$,
where $\mathcal{D}$ is a multimodal clip index and $\mathcal{G}$ is an LCE graph built over the same clips.

We first partition $V$ into temporally ordered 30-second clips $\{v_i\}_{i=1}^{N}$ and caption each visual segment $v_i$ over interval $[s_i,e_i]$. 
Following \cite{DBLP:conf/nips/ZhangJGLLLL25}, we then merge clip-level subject descriptions into a video-level registry and remap recurring entities across captions.
Each record in $\mathcal{D}$ stores the interval, caption, and a fused clip embedding, while the subject registry is retained as video-level metadata.
We then build the LCE graph from the indexed clips and registry, define temporal neighborhoods from clip order, add multimodal-similarity connections, and embed its locator nodes for query-time matching. 
The resulting database integrates complementary retrieval structures: $\mathcal{D}$ supports direct matching of multimodal clip content, while $\mathcal{G}$ enables structured traversal across temporally and semantically related observations.

\noindent\textbf{Multimodal clip index.}
To support flexible combinations of input modalities, we leverage the off-the-shelf omni-modal embedder $f_{\mathrm{mm}}$ (OmniUE-7B \cite{DBLP:journals/corr/abs-2608-27044} in this paper), which maps arbitrary combinations of modalities into a unified embedding space.
Specifically, each visual segment $v_i$ and its associated caption $c_i$ are jointly encoded as $z_i=f_{\mathrm{mm}}([v_i,c_i])\in\mathbb{R}^{d}$. 
Given a text query $x$, we use the same embedder to compute $z_x = f_{\mathrm{mm}}([x])$ and rank clips by the cosine similarity of their embeddings to $z_x$.
This unified embedding space not only enables text-to-visual-text retrieval without post-hoc merging but also naturally accommodates audio when available\footnote{Experiments with audio are discussed in Sec.~\ref{sec:ablation}.}, while preserving visual cues that may be absent from captions and maintaining the language-based retrieval interface required by the agent.

\noindent\textbf{Locator-Concept-Evidence (LCE) graph.}
While a multimodal index alleviates the limitations of text-centric memory, its independent clip-level retrieval makes it difficult to connect recurring entities and events across temporally distant observations.
To enable cross-clip association, we construct an LCE graph $\mathcal{G}=(\mathcal{L}\cup\{v_i\}_{i=1}^{N},\mathcal{E})$, where each locator $l_j\in\mathcal{L}$ represents an entity, event, or visual attribute (e.g., a person or place) identified from clip captions and a video-level subject registry, and each clip $v_i$ retains its caption $c_i$ and temporal interval.
Each edge in $\mathcal{E}$ links a locator to a clip and encodes a \emph{concept}, a concise description of the locator's role in that clip, along with its supporting caption span, termed \emph{evidence}.
In this manner, locators identify what can connect distant observations (e.g., black cap), concepts describe how each locator participates in a clip (e.g., worn by the young man), and evidence grounds that participation in the caption (e.g., The young man in the black cap).
To construct the LCE graph, we employ a VLM to extract recurring locators from clip captions and the video-level subject registry, characterize their clip-specific roles as concepts, and associate each relation with supporting caption evidence.
This structured representation enables the agent to trace shared entities and events across temporally distant clips while retaining the evidence needed for grounded reasoning; the prompt template is provided in Appendix A.

To capture diverse relationships among clips, we define three types of graph neighborhoods:
(1) Temporal neighborhood is computed on demand from clip order within a window of $\delta$ to preserve local continuity.
(2) Associative neighborhood connects clips linked to a common locator to enable the retrieval of recurring entities and events across distant timestamps. 
(3) Multimodal-similarity neighborhood connects clips with similar multimodal representations to recover semantic correspondences that may be missed by lexical matching due to differing captions or omitted visual details.
Moreover, to prevent fragmented associations caused by inconsistent descriptions across clip batches, we further merge locator variants whose embedding similarity exceeds a threshold $\tau$, consolidating semantically equivalent descriptions (e.g., man with a black cap vs. young man wearing a black hat).

\noindent\textbf{Query-conditioned graph retrieval.}
Given an agent-provided query $\hat{x}$, we first retrieve the nearest locators in the embedding space to identify semantically relevant anchor clips, then expand from these anchors to recover temporally, associatively, and multimodally related evidence that complements the direct match.
The retrieved set $\mathcal{R}(\hat{x})$ ranks the anchor, associative, and multimodal-similarity clips by caption similarity to the question.
Neighbors recovered by temporal proximity are summarized together with their anchors, as temporally adjacent clips often capture continuous events whose context may be fragmented and redundant across clip boundaries.
The retrieval output retains timestamps, captions, related locators, selected route concepts, and neighborhood provenance, enabling query-conditioned distillation into timestamped evidence for the agent.
Unlike flat clip retrieval, which relies on local multimodal similarity, graph retrieval exploits temporal and associative neighborhoods to recover evidence distributed across the video.

\subsection{Multimodal Tool Set}
\label{sec:toolkit}
Since long-video questions require diverse forms of evidence, we provide three database retrieval tools, one source-video inspection tool for fine-grained verification, and a terminal action.
Each tool offers a distinct capability and interface, enabling the controller to retrieve broad context, match clips, trace cross-clip relationships, or verify visual details.

\noindent\textbf{\texttt{(1) Video Overview}.}
When broader video context is needed, this tool provides a query-conditioned overview. 
Given a text query $\hat{x}$, it retrieves the top-$k_{\mathrm{ov}}$ clips from $\mathcal{D}$ by embedding similarity, orders their captions temporally, and generates a timestamped summary addressing $\hat{x}$ based on the retrieved captions.
It also returns the video-level subject registry stored for the video, providing a consistent reference to recurring subjects for subsequent retrieval.
\textbf{\texttt{(2) Clip Retrieval}.}
This tool retrieves clips that directly match a query, providing an efficient way to directly identify relevant moments in the video. 
Given a query $\hat{x}$ and a result count $k$ (default 16), it embeds $\hat{x}$, ranks clips in $\mathcal{D}$ by cosine similarity, and returns the top-$k$ timestamped captions in temporal order.
\textbf{\texttt{(3) LCE Retrieval}.}
For questions requiring evidence integration across distant video segments (e.g., counting or temporal reasoning), this tool retrieves $\mathcal{R}(\hat{x})$ by identifying relevant locators and expanding their neighborhoods to recover related evidence.
\textbf{\texttt{(4) Visual Verification}.}
When retrieved evidence requires further examination, this tool allows the agent to directly inspect the source video for fine-grained visual and temporal verification.
Given a text question and candidate timestamp ranges, it uniformly samples a total of $N_F$ frames across all ranges per tool call and submits them to a VLM.
The resulting answer indicates whether the visual evidence supports the hypothesis or remains inconclusive.
\textbf{\texttt{(5) Return Answer}.}
This terminal action ends the retrieval process and generates a final answer grounded in the accumulated evidence.

\subsection{Model Predictive Retrieval Control (MPRC)}
\label{sec:mprc}

Existing video agents typically select and execute one action at a time based on the current decision state $b_t$, leading to myopic decisions that overlook opportunities for more informative future observations. 
Inspired by model predictive control (MPC) used in control planning ~\cite{camacho2007constrained,kouvaritakis2016model}, we formulate LVU as a sequential optimization problem and introduce MPRC, which approximates long-horizon retrieval planning by repeatedly optimizing candidate trajectories over a finite horizon, executing the first action, and replanning based on newly retrieved evidence.
This receding-horizon strategy balances immediate information acquisition against its anticipated downstream utility while adapting retrieval decisions to newly observed evidence.

MPRC extends the conventional reason-then-act loop into four stages: 
(1) \emph{Reason}: Propose candidate trajectories conditioned on the current decision state; 
(2) \emph{Verify}: Estimate the expected answerability and information gain of candidate trajectories using a dynamics model and select the best trajectory; 
(3) \emph{Execute}: Perform only the first action of the selected trajectory; 
and (4) \emph{Replan}: Incorporate the resulting observation into the interaction history and generate new trajectories from the updated decision state.

\noindent\textbf{Candidate trajectory proposal.}
At each step, the agent generates a set $\Pi_t$ of $K$ candidate retrieval trajectories, each comprising $S$ sequential tool calls:
\begin{equation}
    \Pi_t
    =
    \left\{
    \pi_t^k=(a_{t,k}^{1},\ldots,a_{t,k}^{S})
    \right\}_{k=1}^{K},
    \label{eq:proposals}
\end{equation}
where $\pi_t^k$ denotes the $k$-th hypothetical open-loop trajectory conditioned on the current decision state $b_t$, including the history of executed actions and observations.

\noindent\textbf{Retrieval dynamics prediction.}
In conventional MPC, candidate trajectories are evaluated by rolling out a dynamics model to predict their future states. For LVU, however, executing every candidate trajectory against the actual retrieval environment would incur substantial tool-interaction costs.
Thus, we introduce an LLM-based retrieval dynamics model $\mathcal{F}$, inspired by LLM-based world models for planning~\cite{DBLP:conf/emnlp/HaoGMHWWH23,DBLP:journals/corr/abs-2601-08955}, to estimate the qualitative information consequences of candidate retrieval actions without executing them in the real environment.
For each candidate trajectory $\pi_t^k$, $\mathcal{F}$ predicts an information state at each planning step $s$:
\begin{equation}
    \hat{\iota}_{t,k}^{s}
    =
    \left[\mathcal{F}\!\left(b_t,\pi_t^k\right)\right]_{s}
    =
    \left(
    g_{t,k}^{s},
    u_{t,k}^{s},
    \bar{u}_{t,k}^{s}
    \right),
    \label{eq:predicted-state}
\end{equation}
where $g_{t,k}^{s}$ denotes the expected information category (e.g., the country named with the winning film), $u_{t,k}^{s}$ the uncertainty targeted by the action (e.g., which country that film is submitted from), and $\bar{u}_{t,k}^{s}$ the expected remaining uncertainty (e.g., whether the on-screen text confirms it). 
For multi-step trajectories, the predicted information categories and residual uncertainties from preceding steps provide qualitative context for estimating subsequent information consequences. 
Importantly, this process does not propagate imagined observations as established facts; instead, it characterizes which uncertainties a proposed sequence could plausibly address and which would remain unresolved, allowing the verifier to compare the expected evidential coverage of alternative retrieval plans.

\begin{table*}[t]
    \centering
    \caption{Main results on LVU benchmarks (accuracy, \%). Results for VLMs are from~\cite{yin2026videoarm,DBLP:journals/corr/abs-2608-22516}, except for GPT-6.1-Sol, which we evaluate ourselves. \modelname is evaluated without audio input. ``--'': unreported results. We highlight the \textbf{best} and \underline{second-best} results.}
    \vspace{-5pt}
    \label{tab:main-results}
    \setlength{\tabcolsep}{16pt} % Adjust column spacing
    \scalebox{0.75}{
    \begin{tabular}{lcccc}
        \toprule
        Method & EgoSchema (val) & LVBench (w/o sub) & VideoMME (long, w/o sub) & LongVideoBench (val, long) \\
        \midrule
        \rowcolor{gray!10}
        \multicolumn{5}{c}{\textit{Commercial VLMs}} \\
        Gemini-2.5-Pro~\cite{DBLP:journals/corr/abs-2507-06261} & 72.2 & 67.4 & 82.4 & 69.8 \\
        Gemini-3.1-Pro \cite{gemini-31-pro} & 76.4 & 78.2 & 80.3 & 77.0 \\
        GPT-5~\cite{openai-gpt5} & 73.8 & 59.8 & 68.4 & 61.8 \\
        GPT-6.1-Sol~\cite{openai-gpt61} & \underline{82.4} & 74.7 & 82.3 & 77.5 \\
        \midrule

        \rowcolor{gray!10}
        \multicolumn{5}{c}{\textit{Open-Source VLMs}} \\
        mPLUG-Owl3~\cite{DBLP:conf/iclr/Ye0LH0000025} & -- & 43.5 & 50.1 & 59.8 \\
        InternVideo3~\cite{DBLP:journals/corr/abs-2606-12195} & 76.6 & 55.7 & 73.8 & 66.8 \\
        Qwen2.5-VL-72B~\cite{DBLP:journals/corr/abs-2502-13923} & 76.2 & 47.3 & 72.6 & 65.9 \\
        Qwen3-VL-30B-A3B~\cite{DBLP:journals/corr/abs-2511-21631} & -- & 67.7 & 79.2 & -- \\
        LLaVA-OV-2~\cite{DBLP:journals/corr/abs-2605-25979} & -- & 55.5 & 76.3 & 66.9 \\
        AdaReTaKe~\cite{DBLP:conf/acl/WangSZWCN25} & -- & 53.3 & 73.5 & 67.0 \\
        
        \midrule

        \rowcolor{gray!10}
        \multicolumn{5}{c}{\textit{Agentic / Training-Free}} \\
        VideoAgent~\cite{DBLP:conf/eccv/WangZZY24} & 63.2 & 29.3 & 46.4 & -- \\
        VideoTree~\cite{DBLP:conf/cvpr/WangYSYCBB25} & 67.0 & 28.8 & 54.2 & -- \\
        DVD~\cite{DBLP:conf/nips/ZhangJGLLLL25} (o3+GPT-4.1) & 76.6 & 74.2 & 67.3 & 71.6 \\
        
        LensWalk~\cite{DBLP:conf/cvpr/LiLSLCS26} (o3+GPT-4.1) & 74.8 & 66.8 & 70.0 & 69.9 \\
        VideoARM~\cite{yin2026videoarm} (o3+GPT-4.1) & 78.2 & 79.7 & 75.3 & 73.7 \\
        VideoSeek~\cite{DBLP:conf/cvpr/LinWLSWYLLB26} (GPT-5) & -- & 68.4 & 70.1 & 63.5 \\
        WorldMM~\cite{DBLP:conf/cvpr/YeoKYH26} (GPT-5) & -- & 61.9 & 76.6 & -- \\
        MERIT~\cite{DBLP:journals/corr/abs-2608-07663} (GPT-5) & -- & 71.8 & 77.7 & -- \\
        \midrule
        
        \rowcolor{sonyblue!15}
        \multicolumn{5}{c}{\textit{Ours}} \\
        \midrule
        \modelname (o3+GPT-4.1) & 80.0 & \underline{80.6} & \underline{86.9} & \underline{78.4} \\
        \modelname (GPT-6.1-Sol) & \textbf{85.2} & \textbf{81.5} & \textbf{88.4} & \textbf{86.2} \\
        \bottomrule
    \end{tabular}
    }
\end{table*}

\noindent\textbf{Trajectory verifier.}
To evaluate if a proposal could be helpful for the user query, the verifier $\mathcal{V}$ aims to compare the candidate trajectories and their predicted information states for jointly assessing \emph{answerability}, whether the complete trajectory can resolve a question-determining uncertainty, and \emph{information gain}, whether it is expected to provide evidence beyond $\mathcal{H}_{t-1}$:
\begin{equation}
    k^\star =
    \mathcal{V}\!\left(
    b_t,
    \left\{
    \left(\pi_t^k,\hat{\iota}_{t,k}^{1:S}\right)
    \right\}_{k=1}^{K}
    \right),
\end{equation}
where $k^\star$ is the index of the selected trajectory.
The distinction between these two criteria discourages redundant trajectories that repeatedly acquire already available evidence, even when that evidence remains relevant to the question.
Since explicitly defining and assigning rewards to individual steps is challenging in LVU, we employ an LLM as $\mathcal{V}$ to leverage its semantic reasoning capabilities to evaluate candidate trajectories.

\noindent\textbf{Receding-horizon execution.}
MPRC executes the first action of the selected trajectory as $a_t=a_{t,k^\star}^{1}$.
After observing the actual tool output $o_t$, the agent appends $\langle a_t,o_t\rangle$ to its interaction history and replans from the updated state $b_{t+1}$.

\section{Experiments}
\label{sec:experiments}

\subsection{Experimental setup}
\label{sec:protocol}
\noindent\textbf{Benchmarks.}
Following established evaluation protocols, we evaluate \modelname on four challenging LVU benchmarks:
EgoSchema~\cite{DBLP:conf/nips/MangalamAM23} (validation split; 500 questions) for egocentric understanding; LVBench~\cite{DBLP:conf/iccv/WangHHCZQDGHXDT25} (1,549 questions across 103 hour-long videos) and LongVideoBench~\cite{DBLP:conf/nips/WuLCL24} (validation long split; 564 questions across 188 videos of 900--3,600 seconds) for long-range temporal reasoning; and VideoMME~\cite{DBLP:conf/cvpr/FuDLLRZWZSZCLLZ25} (long split; 900 questions across 300 videos averaging 2,466 seconds) for general LVU.
We report accuracy on all benchmarks without video subtitles, denoted as w/o sub.

\begin{table*}[t]
    \centering
    \caption{Ablation study on the impact of each component. Removing LCE also disables the LCE Retrieval tool. T: Text. V: Visual.}
    \label{tab:component-ablation}
    \vspace{-5pt}
    \setlength{\tabcolsep}{17.5pt} % Adjust column spacing
    \scalebox{0.80}{
    \begin{tabular}{c|ccc|cccc}
        \toprule
        & Embedder & Memory & MPRC & EgoSchema & LVBench & VideoMME & LongVideoBench \\
        \midrule
        i & T & $\times$ & $\times$ & 75.2 & 74.7 & 76.0 & 73.5 \\
        ii & T+V & $\times$ & $\times$ & 75.6 & 76.4 & 80.6 & 75.6 \\
        \midrule
        iii & T+V & MERIT~\cite{DBLP:journals/corr/abs-2608-07663} & $\times$ & 76.0 & 76.3 & 81.4 & 77.0 \\
        iv & T+V & LCE & $\times$ & 78.4 & 76.8 & 83.6 & 77.2 \\
        \midrule
        v & T+V & $\times$ & $\checkmark$ & 79.2 & 78.5 & 82.1 & 77.7 \\
        \rowcolor{sonyblue!15}
        vi & T+V & LCE & $\checkmark$ & \textbf{80.0} & \textbf{80.6} & \textbf{86.9} & \textbf{78.4} \\
        \bottomrule
    \end{tabular}
    }
\end{table*}

\noindent\textbf{Implementation details.}
Database captions are built using GPT-5.4-mini \cite{openai-54-mini}.
For fair comparisons, we use GPT-4.1 \cite{openai-41} for the tool model, retrieval dynamics model $\mathcal{F}$, and verifier $\mathcal{V}$, and o3 \cite{openai-o3} as the orchestrator.
We additionally evaluate the performance upper bound of \modelname with GPT-6.1-Sol \cite{openai-gpt61}.
We set $K=3$ trajectories, horizon $S=2$ for MPRC, and step budget $T=15$ following \cite{DBLP:conf/nips/ZhangJGLLLL25}.
LCE retrieval returns up to 32 clips with at most eight anchors per route.
We set $N_F$ as 50, $\delta$ as 2 ($\pm2$ neighboring clips, excluding the anchor), $\tau$ as 0.8, and $k_{\mathrm{ov}}$ as 300.
Each MPRC turn comprises one batched proposal, one batched dynamics prediction, one verification, and the selected tool call; increasing $K$ or $S$ thus expands structured outputs instead of requiring $K\times S$ separate dynamics calls.
All main results are averaged over three random seeds.

\subsection{Main Comparison}
\label{sec:main-results}
Tab. \ref{tab:main-results} summarizes the performance of commercial, open-source, and agentic approaches on EgoSchema, LVBench, VideoMME, and LongVideoBench, which demonstrates that our \modelname consistently outperforms all approaches across all four benchmarks.
While Gemini-3.1-Pro and Gemini-2.5-Pro outperform GPT-5 on most benchmarks, GPT-6.1-Sol achieves the strongest overall performance among commercial and open-source VLMs.
Comparisons between agentic baselines and commercial VLMs reveal that recent advances in commercial VLMs have narrowed or even eliminated the performance gap with existing agentic approaches built upon earlier-generation models.
Nonetheless, \modelname, even when equipped with older base models, consistently surpasses both prior video agents using the same backbone models (o3+GPT-4.1) and more recent commercial VLMs (except for EgoSchema), highlighting the effectiveness of multimodal navigational memory and diverse look-ahead planning in extending the capabilities of existing models.
Furthermore, upgrading \modelname to GPT-6.1-Sol yields an additional \textbf{3.9-point} average improvement in LVU performance over its o3+GPT-4.1 counterpart, with significant improvements on EgoSchema and LongVideoBench, demonstrating the generalizability of our agentic framework across different backbone models as well as the benefits of strong foundation models and the LVU superiority.

\begin{table}[t]
    \centering
    \caption{Analyses of MPRC on LVBench.}
    \label{tab:different-mprc}
    \vspace{-5pt}
    \setlength{\tabcolsep}{19pt} % Adjust column spacing
    \scalebox{0.75}{
    \begin{tabular}{l|cc}
        \toprule
        Configuration & Accuracy & Tool calls/Q \\
        \midrule
        w/o MPRC & 76.8 & 11.1 \\
        \midrule
        $K{=}1,S{=}1$ & 77.4 & 10.0 \\
        $K{=}1,S{=}2$ & 78.9 & 10.3 \\
        $K{=}3,S{=}1$ & 78.8 & 9.5 \\
        \midrule
        \rowcolor{sonyblue!15}
        $K{=}3,S{=}2$ & 80.6 & 12.3 \\
        
        \hspace{1.5em}w/o $\mathcal{F}$ & 79.5 & 11.7 \\
        \hspace{1.5em} Joint $\mathcal{F}$ + $\mathcal{V}$ & 78.9 & 11.4 \\
        \midrule
        $K{=}3,S{=}3$ & 80.2 & 12.8 \\
        $K{=}5,S{=}2$ & 80.1 & 14.0 \\
        \bottomrule
    \end{tabular}
    }
\end{table}

\begin{table}[t]
    \centering
    \caption{Impacts on different base models for the video agent and the tool/MPRC ($\mathcal{F}$ and $\mathcal{V}$) on LVBench.}
    \label{tab:different-llm}
    \vspace{-5pt}
    \setlength{\tabcolsep}{19pt} % Adjust column spacing
    \scalebox{0.75}{
    \begin{tabular}{ll|c}
        \toprule
        Agent & Tool/MPRC & LVBench \\
        \midrule
        \rowcolor{sonyblue!15}
        o3 & GPT-4.1 & 80.6 \\
        & GPT-5.4-mini & 81.0 \\
        & Qwen3-VL-8B & 72.7 \\
        \midrule

        GPT-6.1-Sol & GPT-4.1 & 80.8 \\
        & GPT-5.4-mini & 80.8 \\
        & Qwen3-VL-8B & 74.2 \\
        \rowcolor{sonyblue!15}
        & GPT-6.1-Sol & 81.5 \\
        
        \bottomrule
    \end{tabular}
    }
\end{table}

\begin{table}[t]
    \centering
    \caption{Effects of embedding modality on LVBench and VideoMME that have audio. T: text; V: visual; A: audio.}
    \label{tab:different-modalities}
    \vspace{-5pt}
    \scalebox{0.75}{
    \begin{tabular}{ll|cc}
        \toprule
        Embedder & Modality & LVBench & VideoMME \\
        \midrule
        text-embedding-3-large \cite{openai-text-3-large} & T & 78.6 & 82.1 \\
        \midrule
        OmniUE-7B \cite{DBLP:journals/corr/abs-2608-27044} & T & 78.3 & 82.5 \\
        \rowcolor{sonyblue!15}
        & T+V & 80.6 & 86.9 \\
        & T+V+A & 81.3 & 87.7 \\
        \bottomrule
    \end{tabular}
    }
    \vspace{-5pt}
\end{table}

\subsection{Ablation Study}
\label{sec:ablation}

\noindent\textbf{1) Contributions of each module.}
To analyze the relative contributions of the proposed components and tools, we conduct ablation studies by selectively removing different designs.
As shown in Tab.~\ref{tab:component-ablation}, the performance degradation with text-only embeddings (i) suggests that converting visual information into textual descriptions neglects fine-grained visual cues, highlighting the importance of incorporating visual and textual information into a unified embedding space for effective multimodal retrieval.
In addition, incorporating either LCE (iv) or MPRC (v) independently improves performance across all benchmarks, demonstrating their effectiveness across diverse LVU scenarios, including egocentric understanding and long-term memory reasoning.
Moreover, combining both components (vi) achieves the best performance across all benchmarks, with particularly notable gains on VideoMME, indicating that navigational memory and look-ahead planning complement each other in addressing tasks requiring both fine-grained visual perception and higher-level reasoning.
To verify the effectiveness of our proposed LCE, we replace (iv) with MERIT, which utilizes event, dialogue, object, and summary keys as well as visual information.
The declined performance of (iii) suggests the necessity of maintaining multi-relational memory navigation.

\noindent\textbf{2) Effects of look-ahead strategies in MPRC.}
Tab.~\ref{tab:different-mprc} investigates how the number of candidate trajectories ($K$) and planning horizon ($S$), and the dynamics model ($\mathcal{F}$) affect LVU performance and tool-use efficiency.
Compared with the direct reason-then-act baseline (w/o MPRC), our default configuration ($K=3$, $S=2$) achieves higher accuracy with only one additional tool call per question on average.
Notably, both $K=1$, $S=2$ and $K=3$, $S=1$ outperform w/o MPRC and $K=1$, $S=1$, while requiring a comparable number of tool calls, suggesting that look-ahead planning and candidate diversification can each improve retrieval decisions.
However, increasing either $K$ or $S$ beyond the default setting yields diminishing returns or even degrades performance, indicating the importance of balancing planning complexity and effectiveness.
To isolate the contribution of information-consequence prediction, we further remove the dynamics model $\mathcal{F}$ with $K=3$, $S=2$.
Without $\mathcal{F}$, the verifier directly evaluates the proposed retrieval trajectories (similar to Tree of Thoughts~\cite{DBLP:conf/nips/YaoYZS00N23}), achieving 79.5\% accuracy compared with 76.8\% for the direct reason-then-act baseline. 
Incorporating $\mathcal{F}$ further improves accuracy to 80.6\%, demonstrating the complementary benefits of multi-trajectory planning and explicit information-state prediction for retrieval action selection.
To understand if separating retrieval dynamics prediction from trajectory verification is beneficial, the joint $\mathcal{F}+\mathcal{V}$ variant uses a single LLM call to perform both information-consequence prediction and trajectory selection.
Compared with this joint formulation, the decoupled design achieves higher accuracy, suggesting that explicitly separating information-state estimation from trajectory-level evaluation improves retrieval decision-making.

\begin{figure}[t]
    \centering
    \includegraphics[width=\linewidth]{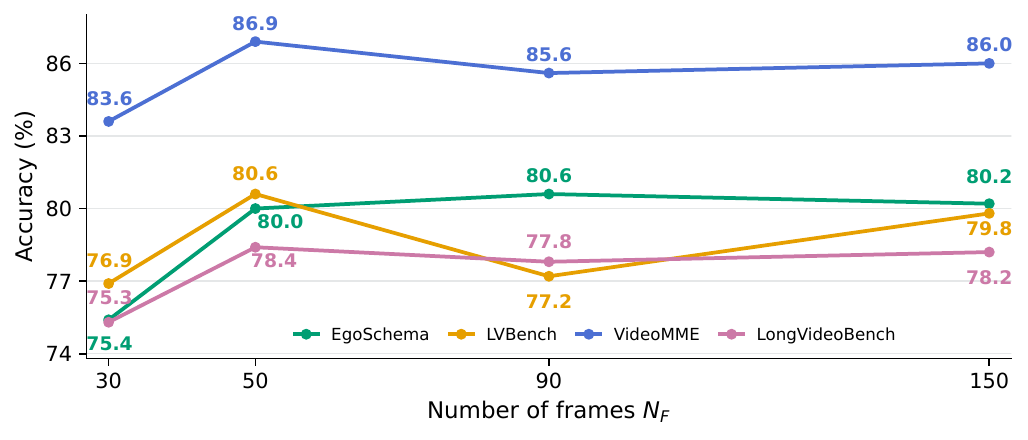}
    \vspace{-24pt}
    \caption{Effect of visual-verification density $N_F$.}
    \label{fig:frame-density}
\end{figure}

\begin{table}[t]
    \centering
    \caption{Effect of the step budget.}
    \label{tab:different-budget}
    \vspace{-6pt}
    \scalebox{0.75}{
    \begin{tabular}{l|cccc}
        \toprule
        Budget & EgoSchema & LVBench & VideoMME & LongVideoBench \\
        \midrule
        5  & 76.4 & 79.0 & 85.5 & 73.8 \\
        10 & 79.8 & 79.8 & 86.2 & 76.5 \\
        \rowcolor{sonyblue!15}
        15 & 80.0 & 80.6 & 86.9 & 78.4 \\
        \bottomrule
    \end{tabular}
    }
    \vspace{-3pt}
\end{table}

\noindent\textbf{3) Effects of embedding modalities.}
Tab.~\ref{tab:different-modalities} assesses the impact of different embedding modalities.
Since LVBench and VideoMME contain synchronized audio, we evaluate different embedding models and modality combinations on these two benchmarks.
While both embedding models achieve comparable performance with text-only embeddings, incorporating visual and audio information consistently improves performance across both benchmarks.
These results highlight the advantages of multimodal memory in a unified space over text-centric representations in preserving rich semantic information that is difficult to fully capture through textual descriptions alone.

\noindent\textbf{4) Effects of base model selection.}
Tab.~\ref{tab:different-llm} compares the performance of \modelname under different base model configurations.
Upgrading the core agent from o3 to GPT-6.1-Sol consistently improves LVU performance, highlighting the importance of strong reasoning capabilities for effective tool use and planning.
While replacing the core agent with GPT-6.1-Sol also improves performance when paired with Qwen3-VL, the remaining gap relative to commercial VLMs suggests potential limitations in handling complex visual understanding tasks.
Nevertheless, employing GPT-6.1-Sol across the core agent, tool use, and MPRC achieves the strongest overall LVU performance, demonstrating the benefits of integrating advanced reasoning and multimodal understanding capabilities within our agentic framework.

\noindent\textbf{5) Effects on visual-verification density.}
Fig.~\ref{fig:frame-density} shows the relations between visual-verification density and LVU performance, showing that performance generally stabilizes as $N_F$ increases from 50 to 150, except for a fluctuation at $N_F=90$ on LVBench.
These results suggest that inspecting additional frames does not necessarily improve evidence identification, as effective visual verification also relies on the agent's ability to reason over the inspected visual evidence.
Considering the trade-off between LVU performance and inference cost, we adopt $N_F=50$ by default.

\noindent\textbf{6) Effects of different step budgets.}
\label{app:cost-analysis}
Tab.~\ref{tab:different-budget} shows how the maximum number of retrieval steps affects \modelname's performance. Increasing the budget from 5 to 15 steps improves accuracy on all four benchmarks by 1.4 to 4.6 percentage points. LongVideoBench shows the largest overall gain (4.6 points) and benefits most from increasing the budget from 10 to 15 steps (1.9 points). 
These results indicate that allowing additional retrieval steps enables the agent to gather more sufficient evidence for answering complex questions; therefore, we use a 15-step budget in our main experiments.

\noindent
More ablation and case studies are depicted in Appendix B.
\section{Conclusion}
\label{sec:conclusion}

In this paper, we introduce \modelname, a novel agentic navigation framework with multimodal navigational memory and predictive control for long-video understanding.
Distinct from existing video agents that primarily rely on text-centric and flat memory, \modelname equips the agent with a unified multimodal clip index and an LCE graph, enabling direct multimodal matching and relational evidence retrieval beyond flat caption-level semantics.
To unlock the agent's receding-horizon planning ability, MPRC consists of a dynamics model predicting the expected results from different trajectories generated by the agent, as well as a verifier to select the trajectory with better utility gain.
Extensive experiments on four LVU benchmarks demonstrate that \modelname not only significantly outperforms existing video agents, but also consistently surpasses the latest commercial models.
We believe \modelname serves as a generic framework for LVU and opens up new opportunities for engaging with unified multimodal embeddings and receding-horizon planning toward more reliable long-form video agents.

{
    \small
    \bibliographystyle{ieeenat_fullname}
    \bibliography{main}
}

% WARNING: do not forget to delete the supplementary pages from your submission

% \appendix
% \newpage

% \clearpage
% \setcounter{page}{1}
% \maketitlesupplementary
\appendix

\section{Prompt Templates}
\label{app:prompts}

\newcommand{\ph}[1]{{\fontencoding{T1}\ttfamily\selectfont\textless\textless #1\textgreater\textgreater}}
\lstdefinestyle{prompt}{
  basicstyle=\scriptsize\fontencoding{T1}\ttfamily,
  breaklines=true,
  breakindent=0pt,
  columns=fullflexible,
  keepspaces=true,
  showstringspaces=false,
  aboveskip=0pt,
  belowskip=0pt,
  literate=
    {<<}{{\textless\textless}}2
    {>>}{{\textgreater\textgreater}}2
}
\tcbset{
  prompt/.style={
    enhanced,
    breakable,
    listing only,
    colback=white,
    colframe=realsonyblue,
    colbacktitle=realsonyblue,
    coltitle=white,
    boxrule=0.4pt,
    arc=1pt,
    left=4pt, right=4pt, top=2pt, bottom=2pt,
    before upper=\setlength{\parindent}{0pt},
    before skip=4pt,
    after skip=6pt,
    fonttitle=\bfseries\footnotesize,
    listing options={style=prompt},
  }
}

% \subsection{Prompt Templates}
% \label{app:prompts}

\subsection{Core Agent}
The core agent receives the following system message and question message at the start of every question.
Subsequent turns extend this conversation with the executed tool calls and their observations.

\begin{tcblisting}{prompt, title={Core Agent: System Prompt}}
You are a helpful assistant who answers multi-step questions by sequentially invoking functions. Follow the THINK -> ACT -> OBSERVE loop:
  * THOUGHT Reason step-by-step about which function to call next.
  * ACTION   Call exactly one function that moves you closer to the final answer.
  * OBSERVATION Summarize the function's output.
Only pass arguments that come verbatim from the user or from earlier function outputs---never invent them. If you are uncertain about code structure or video content, use the available tools to inspect rather than guessing. Plan carefully before each call and reflect on every result. Do not rely solely on blind function calls, as this degrades reasoning quality. Timestamps may be formatted as 'HH:MM:SS' or 'MM:SS'.
\end{tcblisting}

\begin{tcblisting}{prompt, title={Core Agent: Question}}
Carefully read the timestamps and narration in the following script, paying attention to the causal order of events, object details and movements, and people's actions and poses.

Available tools:
* `Video Overview`(query): retrieves the top clips for the query by embedding similarity, orders their captions in time, and returns a timestamped summary of those captions together with the video-level subject registry.
* `Clip Retrieval`(event_description, top_k): embeds the event description, ranks clips by cosine similarity, and returns the top-k timestamped captions in temporal order.
* `LCE Retrieval`(query, question, concepts): matches the query to the nearest locators and expands each match through its temporal, associative, and multimodal-similarity neighborhoods. Anchor, associative, and multimodal-similarity clips are ranked by caption similarity to the question; temporal neighbors are summarized with their anchors. Returns timestamps, captions, related locators, concepts, and neighborhood provenance, distilled into timestamped evidence. Optional concepts restrict which locator-clip relations are expanded.
* `Visual Verification`(question, time_ranges): uniformly samples frames across the given time ranges and returns an answer grounded in those frames.

Based on your observations and tool outputs, provide a concise answer that directly addresses the question. Always end by calling `finish` with the final answer (for multiple choice, the letter only).

Total video length: <<video_length>> seconds.

Question: <<question>>
\end{tcblisting}

\subsection{Tool Use}

\begin{tcblisting}{prompt, title={Tool-Use Proposal}}
MPRC propose step. Propose exactly 3 candidate trajectories. Each trajectory must have exactly 2 steps. Only the FIRST step of the chosen trajectory will be executed this turn; later steps are a short sketch of intended follow-ups.

Available tools (do not invent names; omit injected runtime args):
- LCE Retrieval(query:string(required), question:string(required), concepts:array(optional))
- Visual Verification(question:string(required), time_ranges_hhmmss:array(required))
- Clip Retrieval(event_description:string(required), top_k:integer(optional))
- Video Overview(query:string(required))
- finish(answer:string(required))

Rules:
- Do NOT enumerate multiple-choice options inside any query argument.
- Respond with JSON only, no markdown fences:
{"trajectories":[{"id":1,"steps":[{"tool":"...","arguments":{...}},{"tool":"...","arguments":{...}}],"why":"..."}, ...]}
\end{tcblisting}

\subsection{Retrieval Dynamics Model}
A single call predicts the information states for all steps across candidate trajectories.
Each state comprises three fields: the expected information category $g$, the uncertainty targeted by the action $u$, and the expected residual uncertainty $\bar{u}$, as defined in Eq. 3.
\ph{candidates} denotes the JSON list of proposed trajectories.

\begin{tcblisting}{prompt, title={Retrieval Dynamics Model}}
MPRC predict step (retrieval dynamics model). For EACH candidate trajectory below, and for EACH of its steps in order, predict the expected INFORMATION CONSEQUENCE of executing that retrieval action --- NOT the actual video content. You are estimating what CATEGORY of information the action would surface, not which specific fact it will reveal. For subsequent steps, consider the anticipated information categories and residual uncertainties from preceding steps, without assuming that earlier retrievals necessarily succeed or reveal specific facts.

For each step, fill in:
- likely_retrieved_category: the TYPE of information this action would likely surface (e.g. 'a temporal anchor linking two scenes', 'the identity of the actor performing the action', 'a visual attribute of the object in question') --- never a specific plot fact, name, count, or outcome presented as if already known.
- uncertainty_addressed: which currently-open, question-relevant uncertainty this would plausibly narrow.
- residual_uncertainty: what would still remain unresolved even in the best case for this step.

Hard rule: never state a specific answer-determining fact (which option is correct, a specific name/count/time) as if it is already known.

Candidates:
<<candidates>>

Respond with JSON only, no markdown fences:
{"predictions": [{"id": <id>, "predicted_states": [{"step": 1, "likely_retrieved_category": "<category of info this step would surface, not a specific fact>", "uncertainty_addressed": "<which open uncertainty this narrows>", "residual_uncertainty": "<what remains unresolved>"}, ...one entry per step...]}, ...one entry per candidate...]}
\end{tcblisting}

\subsection{Verifier}
The verifier compares the trajectories and their predicted states under answerability and information gain, and returns the trajectory whose first step is executed.
\ph{candidates} is the JSON list of trajectories together with their predicted states.

\begin{tcblisting}{prompt, title={Trajectory Verifier}}
MPRC verify step. Each candidate includes its predicted future information state (`predicted_states`) from the dynamics model --- these are expectations, not confirmed observations. Only step 1 of the chosen trajectory will run this turn; later steps and predicted_states are the planned rationale, not decoration.

Candidates:
<<candidates>>

Weigh two perspectives together for each candidate:
- Answerability: how likely this trajectory is to resolve the question.
- InformationGain: how much it reduces current uncertainty beyond what is already known (a trajectory that mostly re-confirms known evidence scores low here).

Respond with JSON only, no markdown fences:
{"scores": [{"id": <id>, "answerability": <0-10>, "info_gain": <0-10>}, ...one entry per candidate...], "chosen_id": <id>, "reason": "<one sentence mentioning both perspectives>"}
\end{tcblisting}

\subsection{Additional Prompts}
\label{app:other-prompts}

\noindent\textbf{LCE graph construction.}
\label{app:lce-prompt}
Each window of consecutive clip captions is sent to the extractor below.
The system message is ``You are an information-extraction system for video understanding. You read short video-clip captions and extract grounded (locator, concept) associations. Only output valid JSON. Never invent facts not present in the captions.''
\ph{clips} is the window, with one id and caption per clip.

\begin{tcblisting}{prompt, title={LCE Graph Extraction}}
You are given a window of consecutive video clips, each with an id and its caption. Neighboring captions are provided together so you can resolve pronouns and recurring entities.

For EACH clip, extract its salient associations. A locator is a compact retrieval key (a person, object, place, event, or attribute). A concept is a short phrase describing how that locator participates in this clip. Copy an exact supporting substring into "evidence".

Rules:
- Only use information explicitly present in that clip's caption.
- Prefer concrete entities/objects/events over generic words.
- "locator_type" is one of: person, object, place, event, attribute, time.
- Do NOT paraphrase entities; keep them as they appear.

Clips:
<<clips>>

Output ONLY this JSON (no prose, no code fences):
{
  "clips": [
    {"clip_id": "<id>", "associations": [
      {"locator": "<short locator>", "locator_type": "object", "concept": "<short association>", "evidence": "<exact caption substring>"}
    ]}
  ]
}
\end{tcblisting}

\noindent\textbf{Clip captioning.}
Captions are generated by GPT-5.4-mini from frames sampled at 1\,FPS over 30-second clips.
The caption, merge, and remap calls share the system message ``You are a helpful assistant.''
EgoSchema, Video-MME, LongVideoBench, and LVBench videos use the incremental prompt, which feeds the subject registry accumulated from earlier clips into the current clip.
On the first clip, \ph{historical\_registry} is the text \texttt{(empty; this is the first clip)}; afterward it is the registry JSON.

\begin{tcblisting}{prompt, title={Clip Captioner}}
There are consecutive frames from a video, sampled at 1 frame per second, covering clip_start_time=<<clip_start_time>> to clip_end_time=<<clip_end_time>>.

Historical subject registry (subjects already identified in earlier clips of this video; IDs here are final and must be reused as-is when the same subject appears again):
<<historical_registry>>

Write JSON only.

{
  "new_subject_registry": {
    "<<next_subject_id>>": {
      "name": "<short stable identity; unknown if unnamed>",
      "appearance": ["<hair, face, clothes, distinguishing marks>"],
      "identity": ["<role/kinship/relation shown in this clip>"],
      "first_seen": "<timestamp within this clip>"
    }
  },
  "clip_description": {
    "subjects_present": ["<id from the historical registry or from new_subject_registry above>", "..."],
    "captions": "<time-ordered narration>"
  }
}

Rules:
- Only add to new_subject_registry people/objects that are genuinely NOT already present in the historical subject registry above. If a subject already exists there (matching appearance/identity), reuse its existing "subject_i" id in subjects_present instead of creating a duplicate entry.
- Brand-new subjects must use the next unused ids starting at "<<next_subject_id>>" (then subject_<<next_id_plus_1>>, ...). If there are no new subjects, set new_subject_registry to {}.
- Do not use temporary ids such as "new_subject_1". Every id in subjects_present MUST be a key of the historical registry or of new_subject_registry in this same JSON.
- The top-level object MUST contain "clip_description" with a non-empty "captions" string. Do not omit clip_description. Do not put the narration only at the top level.
- Name each person/object with one specific phrase you would reuse in later clips (not "the man" / "vehicles" if a more specific identity is visible).
- State relations explicitly (who with whom, kinship, where they sit/stand, who acts on whom). After X, Y must be written as sequence, not implied.
- When countable or attributed things are visible, include the number, color, and clothing. Action outcomes that are visible must be stated plainly.
- Narrate in chronological order through the clip. Mention setting/room type and salient props with searchable nouns.
- If several subjects are present, do not refer to them only by pronoun.
\end{tcblisting}

\noindent\textbf{Video overview.}
The tool retrieves the top clips for the agent query and asks the tool model to answer from their captions.
The system message is ``You are a knowledgeable assistant specializing in analyzing video content and providing detailed, insightful answers.''
The tool then returns that answer together with the video-level subject registry.

\begin{tcblisting}{prompt, title={Video Overview}}
Below are descriptions of video clips, each with its corresponding timestamp. Carefully review the sequence of events, the details and movements of objects, and the actions and poses of people. Based on these observations, provide a thorough and specific answer to the following question, referencing key events and timestamps as appropriate.
Question: <<query>>

<<clip_captions>>
\end{tcblisting}

\noindent\textbf{Visual verification.}
The tool samples frames uniformly on the requested time ranges and sends them to the tool model with the question.
The system message is ``You are a helpful assistant to answer questions.''

\begin{tcblisting}{prompt, title={Visual Verification}}
Carefully watch the <<n_frames>> video frames sampled uniformly from the given time range(s). Pay attention to the cause and sequence of events, the detail and movement of objects and the action and pose of persons.

If no relevant content is found within the given time range, return: `Error: Cannot find corresponding result in the given time range.`.
Question: <<question>>
\end{tcblisting}

\noindent\textbf{LCE evidence distillation.}
After LCE retrieval expands a neighborhood, the fetched captions are compressed for the original question.
The system message is ``You compress video-clip evidence for a specific question. Keep only what helps answer it; keep timestamps; be concise and factual.''

\begin{tcblisting}{prompt, title={LCE Evidence Distillation}}
Question: <<question>>

Clip evidence (expanded neighborhood):
<<clips>>

Return a short, timestamped summary of only the query-relevant evidence.
\end{tcblisting}

\section{Additional Experiments}
\label{app:additional-experiments}

\begin{table}[t]
    \centering
    \caption{Effects of clip length on LVU performance.}
    \label{tab:different-db}
    \vspace{-5pt}
    \scalebox{0.75}{
    \begin{tabular}{l|cccc}
        \toprule
        Clip Length & EgoSchema & LVBench & VideoMME & LongVideoBench \\
        \midrule
        5 & 79.6 & 78.2 & 83.5 & 77.4 \\
        \rowcolor{sonyblue!15}
        30 & 80.0 & 80.6 & 86.9 & 78.4 \\        
        \bottomrule
    \end{tabular}
    }
\end{table}

\begin{table}[t]
    \centering
    \caption{Effects of temporal neighborhood radius $\delta$ on LVBench.}
    \label{tab:different-pm}
    
    \scalebox{1}{
    \begin{tabular}{l|c}
        \toprule
        $\delta$ & LVBench \\
        \midrule
        0 & 79.7 \\
        1 & 80.0 \\
        \rowcolor{sonyblue!15}
        2 & 80.6 \\
        4 & 80.4 \\    
        \bottomrule
    \end{tabular}
    }
\end{table}

\begin{table}[t]
    \centering
    \caption{Effects of locator merging threshold on LVBench.}
    \label{tab:different-tau}
    
    \scalebox{1}{
    \begin{tabular}{l|c}
        \toprule
        $\tau$ & LVBench \\
        \midrule
        0.6 & 80.5 \\
        \rowcolor{sonyblue!15}
        0.8 & 80.6 \\
        1.0 (no merging) & 80.1 \\    
        \bottomrule
    \end{tabular}
    }
\end{table}

\begin{table}[t]
    \centering
    \caption{Ablation of predicted information-state components in MPRC on LVBench.}
    \label{tab:different-states}
    
    \scalebox{1}{
    \begin{tabular}{l|c}
        \toprule
        Predicted Components & LVBench \\
        \midrule
        $g$ & 79.8 \\
        $u$ & 79.3 \\
        $\bar{u}$ & 79.9 \\
        $g, u$ & 80.4 \\
        $u, \bar{u}$ & 80.1 \\
        \rowcolor{sonyblue!15}
        $g, u, \bar{u}$ & 80.6 \\    
        \bottomrule
    \end{tabular}
    }
\end{table}

\begin{table}[t]
    \centering
    \caption{Ablation of retrieval neighborhoods and concept-labeled relations in LCE on LVBench.}
    \label{tab:different-lce}
    
    \scalebox{1}{
    \begin{tabular}{l|c}
        \toprule
        LCE Configurations & LVBench \\
        \midrule
        Temporal & 77.9 \\
        Associative & 78.2 \\
        Multimodal & 77.5 \\
        Temporal+Associative & 79.7 \\
        Associative+Multimodal & 78.8 \\
        \midrule
        Full neighbors w/o Concept & 78.4 \\
        \rowcolor{sonyblue!15}
        LCE (Temporal+Associative+Multimodal) & 80.6 \\    
        \bottomrule
    \end{tabular}
    }
\end{table}

% \begin{table}[t]
%     \centering
%     \caption{Different tool set on LVBench.}
%     \label{tab:different-toolset}
%     \scalebox{0.75}{
%     \begin{tabular}{cccc|c}
%         \toprule
%         Tool & & &  & LVBench \\
%         \texttt{Video Overview} & \texttt{Clip Retrieval} & \texttt{LCE Retrieval} & \texttt{Visual Verification} \\
%         \midrule
%          \\
%          & & & & 78.5\\
%         \rowcolor{sonyblue!15}
%          & & & & 80.6 \\
%         \bottomrule
%     \end{tabular}
%     }
% \end{table}

% \subsection{Run-to-Run Variability}
% To assess the statistical significance of \modelname (o3+GPT-4.1), standard deviations of accuracy on EgoSchema, LVBench, VideoMME, and LongVideoBench are 0.33, 0.32, 0.48, and 0.40 percentage points, respectively, demonstrating the consistent robustness of our proposed method.

% \subsection{Token Consumption Analysis}
% To quantitatively demonstrate the computaitonal efficiency of our framework, 

\subsection{Effects of Clip Length on Database Construction}
Tab.~\ref{tab:different-db} compares the LVU performance of databases constructed using different clip lengths.
We observe that using 30-second clips consistently outperforms 5-second clips, suggesting that longer temporal contexts facilitate more coherent database construction by preserving contextual information and relationships among entities, benefiting both the entity registry and LCE.

\subsection{Effects of Temporal Neighborhood Size}
Tab.~\ref{tab:different-pm} investigates the impact of the temporal neighborhood radius $\delta$ in LCE retrieval. Increasing $\delta$ from 1 to 2 improves LVBench accuracy from 80.0\% to 80.6\%, while further expanding it to 4 yields no additional gains (80.4\%). These results suggest that a moderate temporal neighborhood provides sufficient local context for evidence retrieval, whereas broader expansion introduces additional clips with limited benefit. 
% We therefore adopt $\delta=2$ by default.

\subsection{Effects of Locator Merging Threshold}
Tab.~\ref{tab:different-tau} investigates the impact of the locator merging threshold $\tau$ in LCE graph construction. Setting $\tau=0.8$ achieves the highest LVBench accuracy of 80.6\%, compared with 80.5\% at $\tau=0.6$ and 80.1\% without merging ($\tau=1.0$). These results suggest that merging semantically equivalent locators facilitates cross-clip associations, while an overly permissive threshold may conflate distinct entities or concepts. We therefore adopt $\tau=0.8$ by default.

\subsection{Effects of Predicted Information-State Components}
Tab.~\ref{tab:different-states} examines the contributions of the three information-state components predicted by the retrieval dynamics model $\mathcal{F}$: anticipated evidence category ($g$), targeted uncertainty ($u$), and residual uncertainty ($\bar{u}$). Predicting individual components yields 79.3-79.9\% accuracy on LVBench, while combining two components improves performance to 80.1-80.4\%. The complete representation achieves the highest accuracy of 80.6\%, showing that jointly modeling anticipated evidence and both targeted and remaining uncertainties provides complementary information for evaluating retrieval trajectories.

\subsection{Effects of LCE Retrieval Components}
Building on the component-level comparisons in Tab. 2, which evaluate LCE against flat multimodal retrieval and MERIT, we further investigate the contributions of its retrieval neighborhoods and concept-labeled relationships.

\noindent\textbf{Effects of retrieval neighborhoods.}
We first evaluate the temporal, associative, and multimodal-similarity neighborhoods individually and in combination. As shown in Tab.~\ref{tab:different-lce}, each neighborhood independently supports evidence retrieval, achieving 77.9\%, 78.2\%, and 77.5\% accuracy on LVBench, respectively. Combining temporal and associative neighborhoods improves accuracy to 79.7\%, suggesting that local temporal continuity and cross-clip entity associations provide complementary evidence. Further incorporating multimodal-similarity retrieval increases accuracy to 80.6\%, indicating that multimodal correspondences complement temporal and associative navigation. These results demonstrate the benefits of integrating all three neighborhoods for long-video understanding.

\noindent\textbf{Effects of concept-labeled relationships.}
To assess whether the LCE graph provides benefits beyond simple locator-to-clip associations, we introduce a variant that removes the concept labels from graph edges while retaining the same locators, clip associations, and three retrieval neighborhoods. As shown in Tab.~\ref{tab:different-lce}, removing concepts decreases accuracy from 80.6\% to 78.4\%, a 2.2 percentage-point drop. This result suggests that explicitly characterizing how locators participate in individual clips provides useful relational context beyond entity co-occurrence alone, facilitating more effective evidence selection and integration across temporally distant observations.

\subsection{Effects of Replacing the Dynamics Model
with Real-Environment Rollouts}

\begin{table}[ht]
    \centering
    \caption{
    Comparison of trajectory evaluation strategies on LVBench.
    Tool calls and inference time per question are normalized to our default configuration with the dynamics model.
    }
    \label{tab:real-rollout}
    \scalebox{0.8}{
        \begin{tabular}{lccc}
        \toprule
         & Acc. (\%) & Tool calls/Q & Time/Q \\
        \midrule
        w/o MPRC & 76.8 & 0.90x & 0.84x \\
        \midrule
        w/o dynamics model & 79.5 & 0.95x & 0.94x \\
        \rowcolor{sonyblue!15}
        w/ dynamics model (ours) & 80.6 & 1.00x & 1.00x \\
        w/ real-environment rollouts & 81.8 & 2.67x & 1.91x \\
        \bottomrule
        \end{tabular}
    }
\end{table}

To assess the accuracy-efficiency trade-off, we replace dynamics predictions with actual tool outputs from candidate rollouts in MPRC under $K=3$ and $S=2$.
All configurations use the same offline memory, backbone
models, and tool implementations.
Tab.~\ref{tab:real-rollout} shows that real-environment rollouts improve accuracy from 80.6\% to 81.8\%, but require $2.67\times$ as many tool calls and $1.91\times$ the inference time. 
This demonstrates that our dynamics model reduces tool calls by 62.5\% and inference time by 47.6\%, at an accuracy cost of 1.2 points.
Moreover, our default configuration improves accuracy by 3.8 percentage points with approximately 19\% higher online inference time compared with w/o MPRC, demonstrating a favorable accuracy-latency tradeoff.
Compared with removing the dynamics model, it improves accuracy by 1.1 points with only 5.3\% more tool calls and 6.4\% longer inference time. These results support predictive evaluation as an efficient approximation to executing
candidate trajectories.

\subsection{Case Studies}

\begin{figure*}[t]
    \centering
    \includegraphics[width=0.88\linewidth]{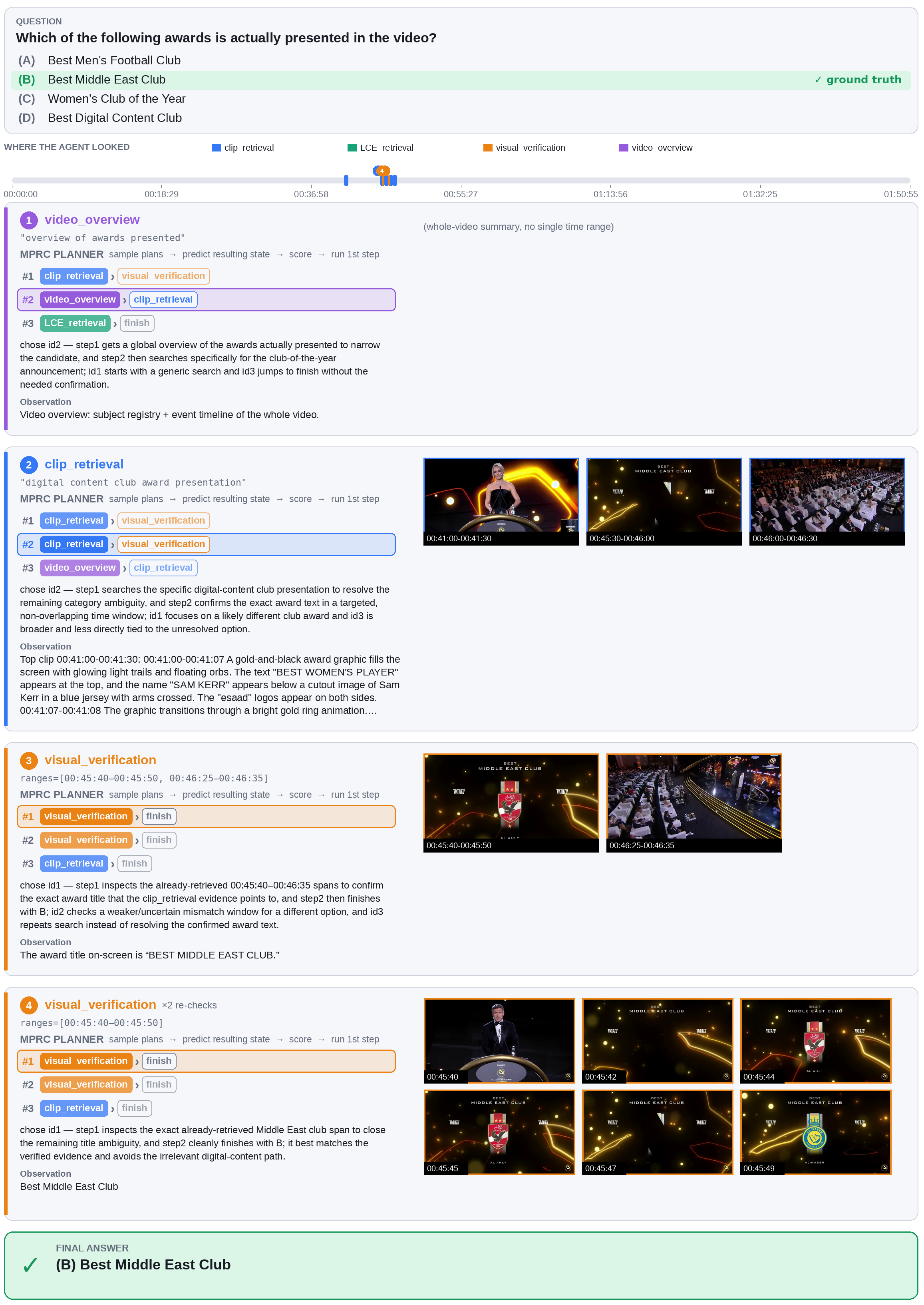}
    \vspace{-10pt}
    \caption{Visualization of the \modelname pipeline on LVBench.}
    % \vspace{-10pt}
    \label{fig:case-lvbench}
\end{figure*}

\begin{figure*}[t]
    \centering
    \includegraphics[width=0.88\linewidth]{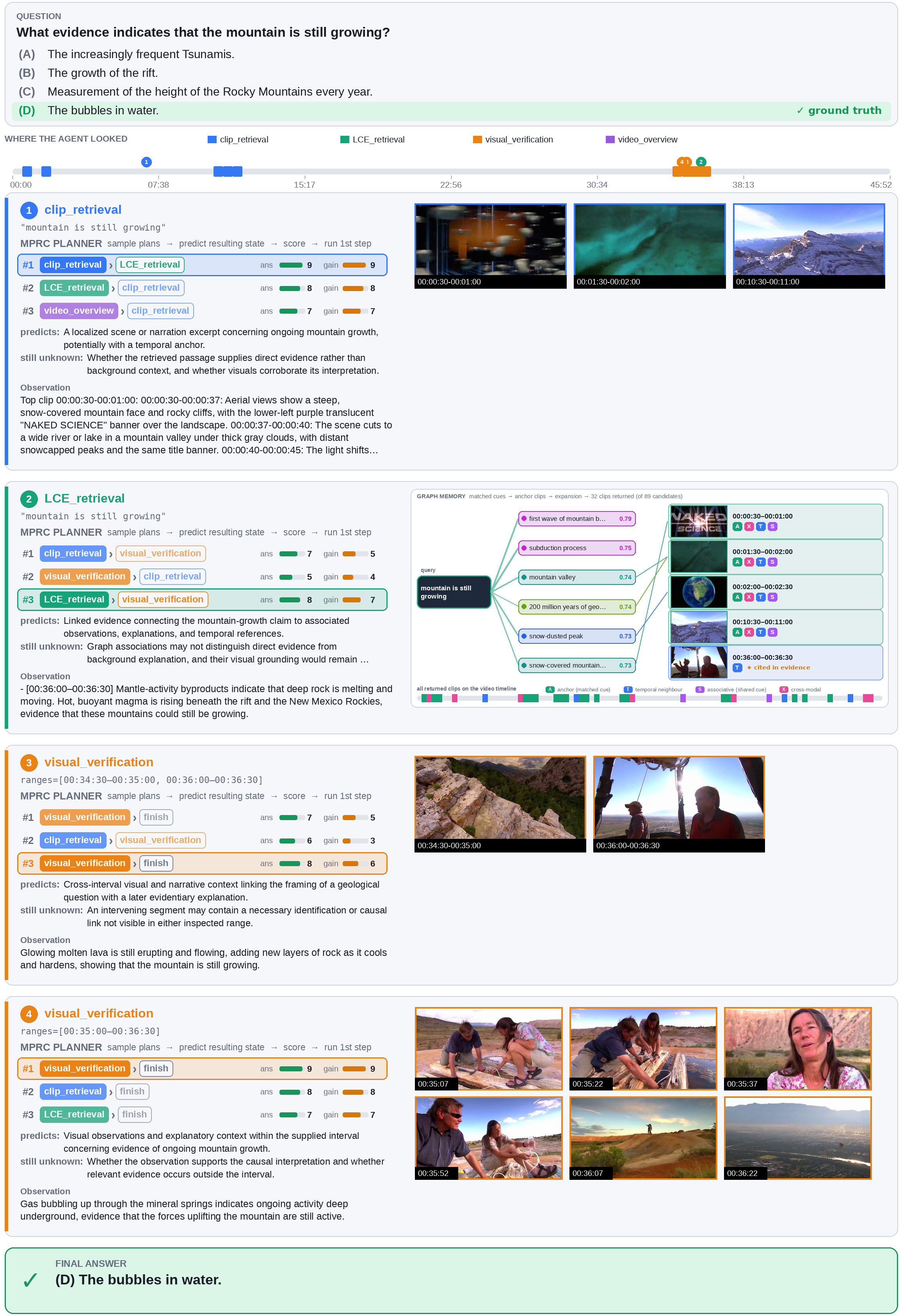}
    \vspace{-10pt}
    \caption{Visualization of the \modelname pipeline on VideoMME.}
    % \vspace{-10pt}
    \label{fig:case-videomme}
\end{figure*}

\begin{figure*}[t]
    \centering
    \includegraphics[width=0.92\linewidth]{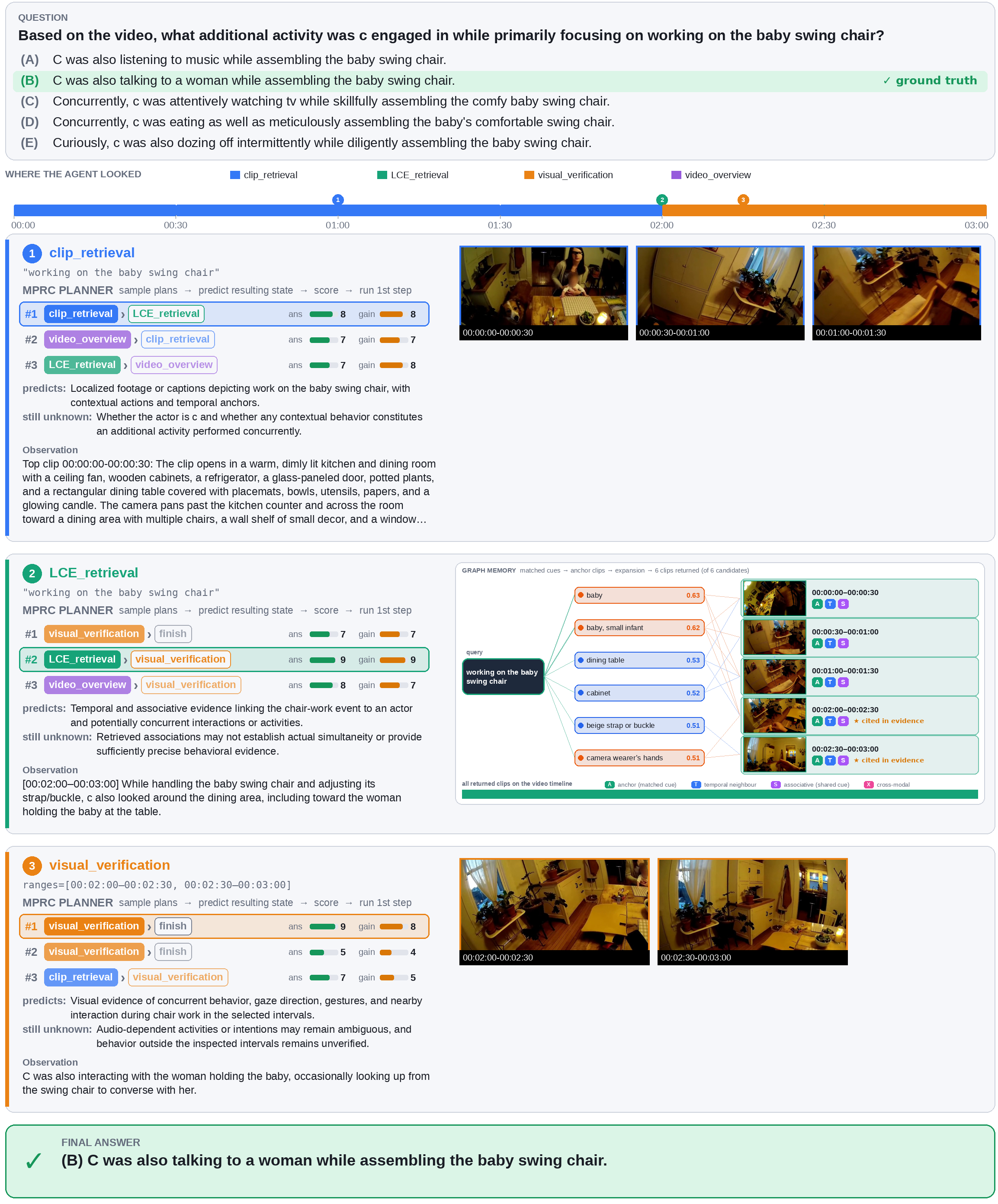}
    \vspace{-10pt}
    \caption{Visualization of the \modelname pipeline on EgoSchema.}
    % \vspace{-10pt}
    \label{fig:case-egoschema}
\end{figure*}

\begin{figure*}[t]
    \centering
    \includegraphics[width=0.92\linewidth]{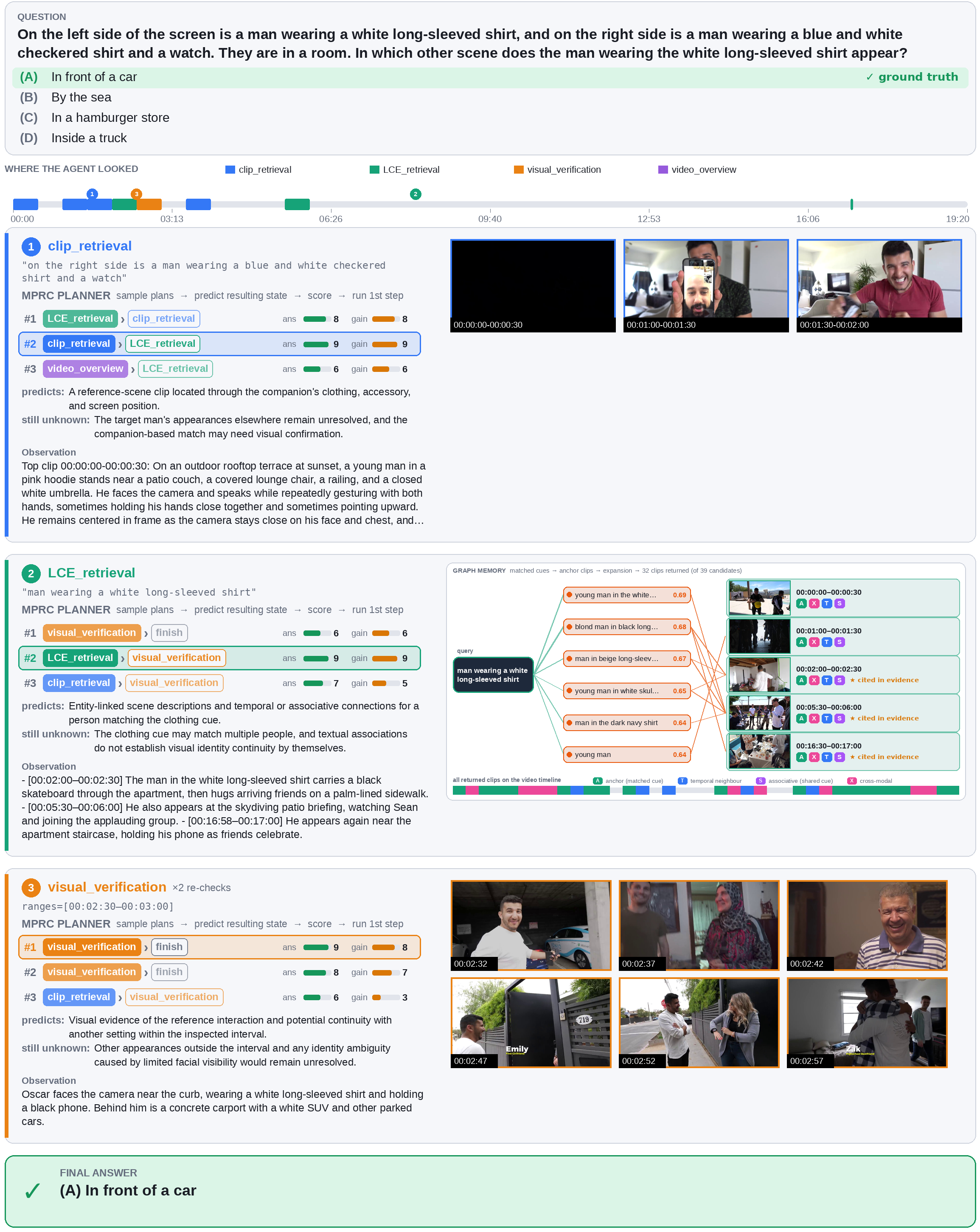}
    \vspace{-10pt}
    \caption{Visualization of the \modelname pipeline on LongVideoBench.}
    % \vspace{-10pt}
    \label{fig:case-longvideobench}
\end{figure*}

Figs.~\ref{fig:case-lvbench}-\ref{fig:case-longvideobench} illustrate how \modelname retrieves evidence, selects tools, and refines its reasoning across different benchmarks.
These cases highlight the interplay between memory navigation and iterative planning in addressing diverse video questions.
For clarity and readability, we present representative cases with shorter reasoning traces, while additional qualitative results featuring more extensive reasoning trajectories and higher-resolution visualizations are provided in the project page.

\end{document}